\documentclass[runningheads]{llncs}

\usepackage[T1]{fontenc}
\usepackage{graphicx}
\usepackage{booktabs}
\usepackage{amsmath,amssymb}
\usepackage[protrusion=true,expansion=false]{microtype}
\usepackage{xcolor}
\PassOptionsToPackage{hyphens}{url}
\usepackage{hyperref}
\usepackage{multirow}
\usepackage{array}
\usepackage{tabularx}
\usepackage{xspace}

\makeatletter
\renewcommand\section{\@startsection{section}{1}{\z@}%
                                  {-10\p@ \@plus -2\p@ \@minus -2\p@}%
                                  {6\p@ \@plus 2\p@ \@minus 2\p@}%
                                  {\normalfont\large\bfseries\boldmath
                                   \rightskip=\z@ \@plus 8em\pretolerance=10000 }}
\renewcommand\subsection{\@startsection{subsection}{2}{\z@}%
                                     {-8\p@ \@plus -2\p@ \@minus -2\p@}%
                                     {4\p@ \@plus 2\p@ \@minus 2\p@}%
                                     {\normalfont\normalsize\bfseries\boldmath
                                      \rightskip=\z@ \@plus 8em\pretolerance=10000 }}
\renewcommand\paragraph{\@startsection{paragraph}{4}{\z@}%
                                    {-6\p@ \@plus -2\p@ \@minus -2\p@}%
                                    {-0.4em \@plus -0.1em \@minus -0.1em}%
                                    {\normalfont\normalsize\itshape}}
\renewcommand\subsubsection{\@startsection{subsubsection}{3}{\z@}%
                                    {-6\p@ \@plus -2\p@ \@minus -2\p@}%
                                    {3\p@ \@plus 1\p@ \@minus 1\p@}%
                                    {\normalfont\normalsize\bfseries\boldmath}}
\makeatother

\let\oldthebibliography\thebibliography
\renewcommand{\thebibliography}[1]{\oldthebibliography{#1}\setlength{\itemsep}{-2pt plus 0.5pt}\setlength{\parsep}{0pt}}

\makeatletter
\def\@listi{\leftmargin\leftmargini \topsep 2\p@ \parsep 0\p@ \itemsep 1\p@}
\let\@listI\@listi
\def\@listii{\leftmargin\leftmarginii \topsep 1\p@ \parsep 0\p@ \itemsep 1\p@}
\makeatother

\newif\ifarxiv
\arxivtrue
\newcommand\blfootnote[1]{%
  \begingroup
  \renewcommand\thefootnote{}\footnote{#1}%
  \addtocounter{footnote}{-1}%
  \endgroup
}

\begin{document}

\title{Output Type Before Quality: A Standards-Derived XAI Admissibility Rubric for Autonomous-Driving Safety}
\titlerunning{Output Type Before Quality}

\author{Abhinaw Priyadershi\inst{1}\orcidID{0009-0007-7120-7972}\thanks{Corresponding author.} \and
Mandar Pitale\inst{1} \and
Jelena Frtunikj\inst{2}\orcidID{0009-0002-5559-1672} \and
Maria Spence\inst{1}}

\authorrunning{A. Priyadershi et al.}

\institute{NVIDIA Corporation, Santa Clara, CA, USA\\
\email{\{apriyadershi,mpitale,mspence\}@nvidia.com} \and
NVIDIA GmbH, Munich, Germany\\
\email{jfrtunikj@nvidia.com}}

\maketitle

\ifarxiv
\blfootnote{This version of the contribution has been accepted for publication, after peer review, but is not the Version of Record and does not reflect post-acceptance improvements or any corrections. It will appear in the proceedings of the SAFECOMP 2026 Workshops (SASSUR), published by Springer in Lecture Notes in Computer Science; the Version of Record will be available online (DOI to be assigned). Use of this Accepted Version is subject to the publisher's Accepted Manuscript terms of use: \url{https://www.springernature.com/gp/open-research/policies/accepted-manuscript-terms}.}
\fi

\begin{abstract}
% SASSUR framing (primary submission target). For WAISE, swap the opening
% sentence to emphasise "AI safety engineering" rather than "assurance case".
Safety standards for ML-based autonomous driving specify the \emph{kind} of evidence an assurance case must contain (directed cause-and-effect chains, quantified interventional effects, named root-cause variables), yet the XAI literature is organised by output type and technique family (saliency maps, feature attribution, counterfactuals, causal graphs, language traces). SHAP, the most-recommended ADS XAI method, returns a ranked feature list that no implementation effort can convert into a directed chain (Fig.~\ref{fig:hero}). We name this mismatch the \emph{evidence-type gap}.

From four publications (AMLAS guidance, ISO~21448, ISO/PAS~8800, ISO~26262) we derive 19 testable evidentiary criteria across seven lifecycle stages with representative clause-cited derivations, and score six XAI method classes structurally: does the method's output \emph{type} contain what the criterion requires? The theoretical anchor is Pearl's causal hierarchy: rung-1 methods cannot answer rung-2 questions.

Causal XAI emerges as structurally required to satisfy the derived criteria at three stages: hazard identification ($+62\%$ rubric gap), incident investigation ($+50\%$), and data management ($+50\%$); the verdict set is stable across thresholds $T \in (0\%, 50\%]$ and survives a worst-case one-step cell flip down to $T = 33\%$. At the remaining four stages, correlational or language-based methods are comparable or sufficient. The rubric identifies structural \emph{admissibility} (necessary but not sufficient for compliance): an admissible method's specific output content may still be wrong, and validating that fidelity (the edges a fitted SCM produces, the cause a trace names) is the open assurance challenge. A single-VLA proof of concept on $1{,}996$ real-world driving clips ($79{,}840$ rows, ten splits) is consistent with each method's observed output type matching its rubric prediction. XAI method selection for ADS safety assurance should be driven by lifecycle-stage evidence demand, not by method popularity.

\end{abstract}

%% ─────────────────────────────────────────────────────────────────
\section{Introduction}
A safety engineer opens ISO/PAS~8800~Cl.~6.7.1 and reads: \emph{``the AI triggering condition activates a fault or a functional insufficiency, resulting in an AI error [\ldots] the AI error of an AI component can propagate through the AI system.''} Cl.~6.7.1 defines this cause-and-effect chain; Cl.~13.3.1 to Cl.~13.3.3 normatively require safety analysis to identify the AI errors, faults, and functional insufficiencies it describes.

\begin{figure}[tb]
\centering
\includegraphics[width=0.95\columnwidth]{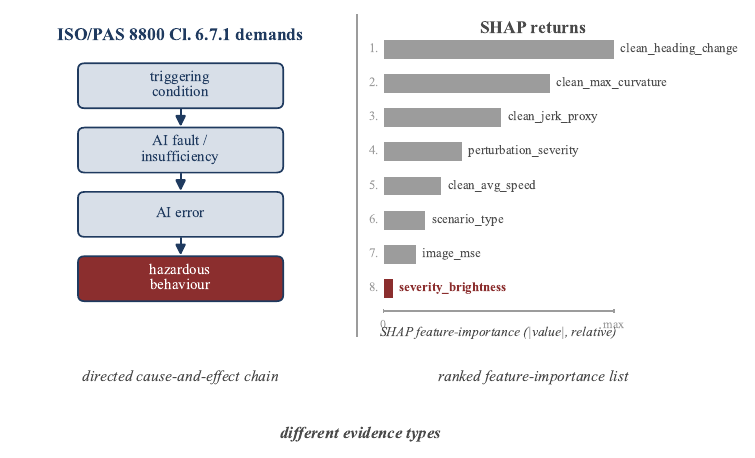}
\caption{\textbf{The evidence-type gap.} A safety standard demands one evidence \emph{type}; a popular XAI method produces another. Left: ISO/PAS~8800~Cl.~6.7.1 defines a directed cause-and-effect chain from triggering condition to hazardous behaviour that Cl.~13.3.1--13.3.3 normatively binds to safety-analysis evidence. Right: SHAP returns a ranked feature-importance list; in a silent-failure example from Section~\ref{sec:validation}, the actual cause (\texttt{severity\_brightness}, in red) is ranked eighth, below seven trajectory-shape and scenario features. The mismatch is structural: the rubric of Section~\ref{sec:gap} predicts it; no improvement to SHAP quality converts a list into a chain. This is a category check, not a comparison of method quality.}
\label{fig:hero}
\end{figure}

The engineer turns to the XAI literature. Kuznietsov et al.~\cite{kuznietsov2024} surveyed 84 ADS XAI papers; the most frequent method is SHAP~\cite{lundberg2017}. TreeSHAP applied to a trajectory-prediction model returns a ranked list of feature importances with ``perturbation\_severity\_norm'' at rank~1. The standard asked for a directed chain. The method returned a ranking. The two are not the same kind of object (Fig.~\ref{fig:hero}). The mismatch is structural, not implementational: Shapley attribution is an associational operation~\cite{pearl2009}, and a ranked list has no edges regardless of how well it is fit.

We formalise when such mismatches occur and identify the lifecycle stages at which they are structural rather than contingent. We make three contributions. (1)~\textbf{A clause-derived rubric:} 19 testable evidentiary criteria across seven ADS lifecycle stages, derived from four publications with representative clause-cited derivations (Table~\ref{tab:criteria_stages}, \S2.4) and a reproducible \textbf{S}/\textbf{P}/\textbf{F} scoring rule (Section~\ref{sec:standards}). (2)~\textbf{Stage-level verdicts under structural scoring:} best-causal (SCM, Causal Trace) vs best-non-causal (SHAP, GradCAM, Counterfactual, CoC) scores per stage, stress-tested under any single S/P/F cell changed one notch, identify causal XAI as structurally required to satisfy the derived criteria at hazard identification ($+62\%$), incident investigation ($+50\%$), and data management ($+50\%$); see Section~\ref{sec:gap}. (3)~\textbf{A single-VLA proof of concept:} on $1{,}996$ real-world driving clips under seven perturbation types at multiple severities, each tested method's observed output type matches its rubric prediction (Section~\ref{sec:validation}). The construction is intended to be domain-general: the same procedure applies to any standard and any method catalogue. Section~\ref{sec:mapping} situates current ADS XAI practice via two recent surveys; Section~\ref{sec:discussion} positions the rubric for XAI method selection in ADS safety assurance and outlines a two-panel validation protocol for follow-up work.

We call the resulting heterogeneity the \emph{evidence-type gap}: the distance between what a lifecycle stage requires and what a given XAI method's output format can in principle provide. At the three causal-necessary stages, correlational methods fall on the wrong side of this gap; at runtime monitoring, language-based chain-of-causation (CoC) traces and SCM-based anomaly signals tie, neither closing the stage uniquely. The gap is not a claim about which method is best, but about which methods are structurally admissible for a given evidentiary demand.

The paper's central claim rests on the structural rubric; the empirical pilot illustrates it on one system. Three scope limits apply: only 3 of 6 methods are empirically touched, 4 of 7 perturbation types have no causal path at $\alpha=0.01$ in the fitted SCM, and learned diagnosis from downstream signals is at chance ($\sim$$30\%$, vs.\ $30.8\%$ majority baseline; Section~\ref{sec:discussion}) in our benchmark.

%% ─────────────────────────────────────────────────────────────────
\section{Evidentiary Demands per Lifecycle Stage}
\label{sec:standards}
\subsection{Scope and Method}

We systematically extracted explainability-related requirements from four published safety standards applicable to ML-based ADS components. For each standard, we identified clauses that explicitly or implicitly require explanatory evidence, tagged each to a lifecycle stage, and classified demand strength as Hard, Soft, or Gap.

The four publications analysed are: \textbf{AMLAS}~\cite{amlas2021}, a six-stage ML safety assurance guidance; \textbf{ISO~21448:2022}~\cite{iso21448}, which addresses safety of intended functionality and the known/unknown scenario framework; \textbf{ISO/PAS~8800:2024}~\cite{isopas8800}, which codifies AI explainability as a safety-related property (Annex~D) and provides a cause-and-effect chain model (Cl.~6.7) bound to normative safety analysis (Cl.~13.3.1--13.3.3), with causal Bayesian networks (Cl.~13.4.3, Table~13-1) and counterfactual analysis (Cl.~9.5.4) listed as analysis techniques; and \textbf{ISO~26262:2018}~\cite{iso26262}, integrated for its V-model lifecycle structure.

\paragraph{Interpretive choices.} Safety standards describe required outcomes without prescribing methods. Converting these into testable XAI criteria requires interpretive judgment. We make two choices readers should be aware of. First, when a standard requires ``causal'' evidence (e.g., ISO/PAS~8800 Cl.~13.3.1--13.3.3's normative cause-and-effect-chain analysis), we interpret this as requiring Pearl's interventional semantics (do-calculus, counterfactuals; distinct from the engineering ``well-reasoned account'' sense), not merely associational evidence labelled as causal. Second, when a standard requires ``identification of insufficiencies'' (e.g., Cl.~11.1(b) for data management), we interpret this as including identification of which triggering conditions are underrepresented. Alternative readings would produce more method-neutral criteria.

\subsection{Lifecycle Stages}

We synthesise the safety lifecycle into seven stages by combining the V-model (ISO~26262), AMLAS, and the SOTIF analysis flow, with 19 criteria distributed across them: (1)~Hazard Identification (H1--H4), (2)~Safety Requirements Definition (R1, R2), (3)~Data Management (D1, D2, D3), (4)~Model Development (M1), (5)~Model Verification (V1, V2), (6)~Runtime Monitoring (RT1--RT3), (7)~Incident Investigation (I1--I4). Stages 1 to 5 correspond to AMLAS S1 to S5 (with stage~1 broadened from AMLAS's ``scoping'' to include hazard identification per ISO~26262 Part~3 Cl.~6 and ISO~21448 Cl.~6). AMLAS S6 covers both runtime and post-incident aspects; we split it into stages~6 and~7 because the XAI evidence types differ: runtime requires real-time anomaly signals (RT1--RT3), incident investigation requires post-hoc fault attribution (I1--I4).

\subsection{Synthesised Evidentiary Demands}

We illustrate the derivation with three representative stages below.

\paragraph{Worked example.}
Cl.~6.7.1 defines the cause-and-effect chain quoted in the introduction, which Cl.~13.3.3 normatively binds to safety analysis (identifying the AI errors, faults, and functional insufficiencies it describes). We formalise the implied \emph{directed} path from input condition to hazardous behaviour as criterion H1: ``output contains a directed causal path from input condition to hazardous behaviour.'' A method scores \textbf{S} if its output type inherently contains directed edges (SCM: a DAG is directed by definition), \textbf{P} if it implies but does not guarantee direction (counterfactual: one path at a time), \textbf{F} if its output type structurally cannot represent direction (SHAP: a ranked list has no edges). We applied the same extract-formalise-score process to all 19 criteria.

Table~\ref{tab:criteria_stages} gives the 11 criteria for three representative stages: hazard identification (H1 to H4), runtime monitoring (RT1 to RT3), and incident investigation (I1 to I4). For H1, H4, I1, I3, I4 the criterion wording tracks the clause text directly; H2, H3, I2 formalise evidentiary demands implied by the clauses (risk/effect evaluation, Bayesian-network conditional independence, root-cause-vs.-symptom distinction) rather than verbatim requirements.\footnote{H2 reads Cl.~13.4.2 numerically; H3 generalises Table~13-1's Bayesian-network conditional-independence advantage.}

\begin{table}[t]
  \centering
  \caption{\textbf{Criteria at three representative stages.} Hazard identification, runtime monitoring, and incident investigation. The remaining stages (R1, R2, D1, M1, V1, V2) follow the same extract-formalise-score procedure. Informative ISO/PAS~8800 annexes also inform RT1, RT2 (G.3.2.1, G.3.3.3) and I1, I2 (Annex~F; G.5.2.2).}
  \label{tab:criteria_stages}
  \renewcommand{\arraystretch}{1.05}
  \begin{tabularx}{\columnwidth}{@{}clX@{}}
    \toprule
    ID & Criterion & Source clause \\
    \midrule
    \multicolumn{3}{@{}l}{\emph{Hazard Identification}} \\
    H1 & Directed cause-and-effect chain & AMLAS S1; ISO/PAS~8800 Cl.~6.7.1, Cl.~13.3.1--13.3.3 \\
    H2 & Quantified effect size & ISO~21448 Cl.~6; ISO/PAS~8800 Cl.~13.4.2 \\
    H3 & Confounding control & ISO/PAS~8800 Cl.~13.4.3 Table~13-1 \\
    H4 & Triggering-condition enumeration & ISO~21448 Cl.~7 (7.1a, 7.3) \\
    \midrule
    \multicolumn{3}{@{}l}{\emph{Runtime Monitoring}} \\
    RT1 & Runtime-feasible signal & AMLAS S6; ISO/PAS~8800 Cl.~14.6.2 \\
    RT2 & Anomaly detection capability & AMLAS S6; ISO/PAS~8800 Cl.~14.6.2 \\
    RT3 & Continuously-updated safety indicator & ISO/PAS~8800 Cl.~14.5, 14.6.2, 14.8.1 \\
    \midrule
    \multicolumn{3}{@{}l}{\emph{Incident Investigation}} \\
    I1 & Fault localisation & AMLAS S6; ISO/PAS~8800 Cl.~14.8 \\
    I2 & Root-cause vs.\ symptom distinction & ISO/PAS~8800 Cl.~6.7.2, 13.4.2 \\
    I3 & Known/unknown trigger classification & ISO~21448 Cl.~13, 4.2.2 \\
    I4 & Assurance-case update support & ISO/PAS~8800 Cl.~14.5 \\
    \bottomrule
  \end{tabularx}
\end{table}

\subsection{Summary}

Together, the 19 criteria distribute across seven stages. Hazard identification and incident investigation have the strongest direct support for causal structure; runtime monitoring requires anomaly signals satisfiable by either causal or language-based methods. At data management, D1 covers dataset-insufficiency identification and the \emph{Completeness} property (ISO/PAS~8800 Cl.~11.1(b), Cl.~11.4.3.2 Table~11-1); D2 covers \emph{Completeness} and \emph{Representativeness} under Cl.~11.3.6 and Table~11-1; D3 links ISO~21448 Cl.~7.3.1 trigger-and-insufficiency traceability to ISO/PAS~8800 Cl.~11.3.7 and Cl.~11.4.4 Table~11-3 dataset-requirement traceability.

%% ─────────────────────────────────────────────────────────────────
\section{XAI Method Inventory and Lifecycle Mapping}
\label{sec:mapping}
\subsection{Method Selection}

We evaluate six XAI methods spanning four categories. \emph{Correlational}: SHAP~\cite{lundberg2017} and GradCAM~\cite{selvaraju2017}. \emph{Example-based}: counterfactual explanations~\cite{wachter2018}. \emph{Causal}: structural causal models (SCM)~\cite{pearl2009} with PC~\cite{spirtes2000} / GES~\cite{chickering2002} graph discovery, and causal tracing~\cite{peters2017}. \emph{Language-based}: chain-of-causation (CoC) reasoning traces produced by VLA models at inference time~\cite{priyadershi2026}, defined as structured $\langle\text{action}, \text{target}, \text{reason}\rangle$ natural-language fields emitted ante-hoc (i.e., as part of the model's native inference output), with no graph structure, fixed training-time vocabulary, observation-level scope, and rung-1 (associational) semantics.

\subsection{Current Usage in Literature}
\label{sec:usage}

Two recent surveys of AV XAI provide a coverage map: Kuznietsov et al.~\cite{kuznietsov2024} (a systematic review of 84 papers, Tables~IV to VII) and Atakishiyev et al.~\cite{atakishiyev2024} (a comprehensive overview and field guide). SHAP, GradCAM, and attention methods dominate \emph{model development} and \emph{verification}; counterfactual methods appear primarily in verification; structural causal models and causal tracing appear at no stage in either corpus. At the three causal-required stages of Section~\ref{sec:gap} (hazard identification, data management, incident investigation), the reported coverage is sparse and confined to correlational methods. The distribution is inferred from the surveys' categorisations; the structural-gap claim rests on the qualitative absence of causal methods, not exact counts.

\subsection{Gap Identification}

Combining Section~\ref{sec:standards}'s evidentiary demands with this usage map exposes a structural mismatch: at hazard identification, data management, and incident investigation the demanded evidence types (directed causal paths, quantified effects, root-cause distinction) are precisely those correlational methods cannot produce, whereas runtime monitoring admits any inference-time anomaly signal and so does not single out a method class. Section~\ref{sec:gap} converts this mismatch into stage-level verdicts using the structural scoring rubric.

%% ─────────────────────────────────────────────────────────────────
\section{Structural Gap Analysis}
\label{sec:gap}
Across the $6 \times 19$ matrix, SCM is the only column with zero failures ($16$~\textbf{S} $\cdot 3$~\textbf{P} $\cdot 0$~\textbf{F}); GradCAM has the most ($2$~\textbf{S} $\cdot 3$~\textbf{P} $\cdot 14$~\textbf{F}). Under the scoring rule defined below, three lifecycle stages emerge as causal-necessary, robust to threshold choice and single-cell flips.

\subsection{Scoring Rubric}

Each method is scored structurally against all 19 criteria: we ask whether the method's output type can in principle satisfy the criterion, regardless of implementation quality. Each pair receives \textbf{S}~(Satisfies), \textbf{P}~(Partial), or \textbf{F}~(Fails). Fig.~\ref{fig:rubric_heatmap} presents the full $6 \times 19$ matrix.

The theoretical basis for ``structural'' is Pearl's causal hierarchy~\cite{pearl2009}: methods operating at the associational rung cannot answer interventional or counterfactual questions regardless of engineering effort. SHAP is an associational attribution method built from Shapley decomposition over a chosen value function~\cite{lundberg2017}, so its output is a rung-1 quantity and cannot produce $\mathbb{E}[Y \mid do(X=x)]$ (rung-2) or $Y_{x}$ (rung-3). Criteria H2 (quantified effect) and I2 (root cause vs.\ symptom) are rung-2 questions; SHAP, GradCAM, and CoC traces all operate at rung~1. The limitation is in the mathematics, not the implementation.

\paragraph{Numeric aggregation.}
We map the ordinal verdicts to $\mathrm{S}=2$, $\mathrm{P}=1$, $\mathrm{F}=0$, so a Partial counts as half a Satisfies. For a lifecycle stage with criterion set $C$, a method $m$ receives stage score $\sigma(m)=\sum_{c \in C}\mathrm{score}(c,m)$, with maximum $2|C|$. Writing $\mathcal{K}$ for the two causal methods (SCM, causal trace) and $\mathcal{N}$ for the four non-causal methods, the rubric gap at a stage is
\[
  \Delta \;=\; \frac{\max_{m \in \mathcal{K}}\sigma(m)\;-\;\max_{m \in \mathcal{N}}\sigma(m)}{2|C|},
\]
and a stage is causal-necessary when $\Delta \geq 30\%$. As a worked example, hazard identification (H1 to H4) gives SCM \textbf{S} on all four ($\sigma=8$ of $8$) against a best non-causal score of $3$ (counterfactual), so $\Delta=(8-3)/8=+62.5\%$; incident investigation and data management both give $+50\%$, and the remaining four stages give $\Delta \leq 0$.

\subsection{Results}

\begin{figure}[t]
\centering
\includegraphics[width=0.85\columnwidth]{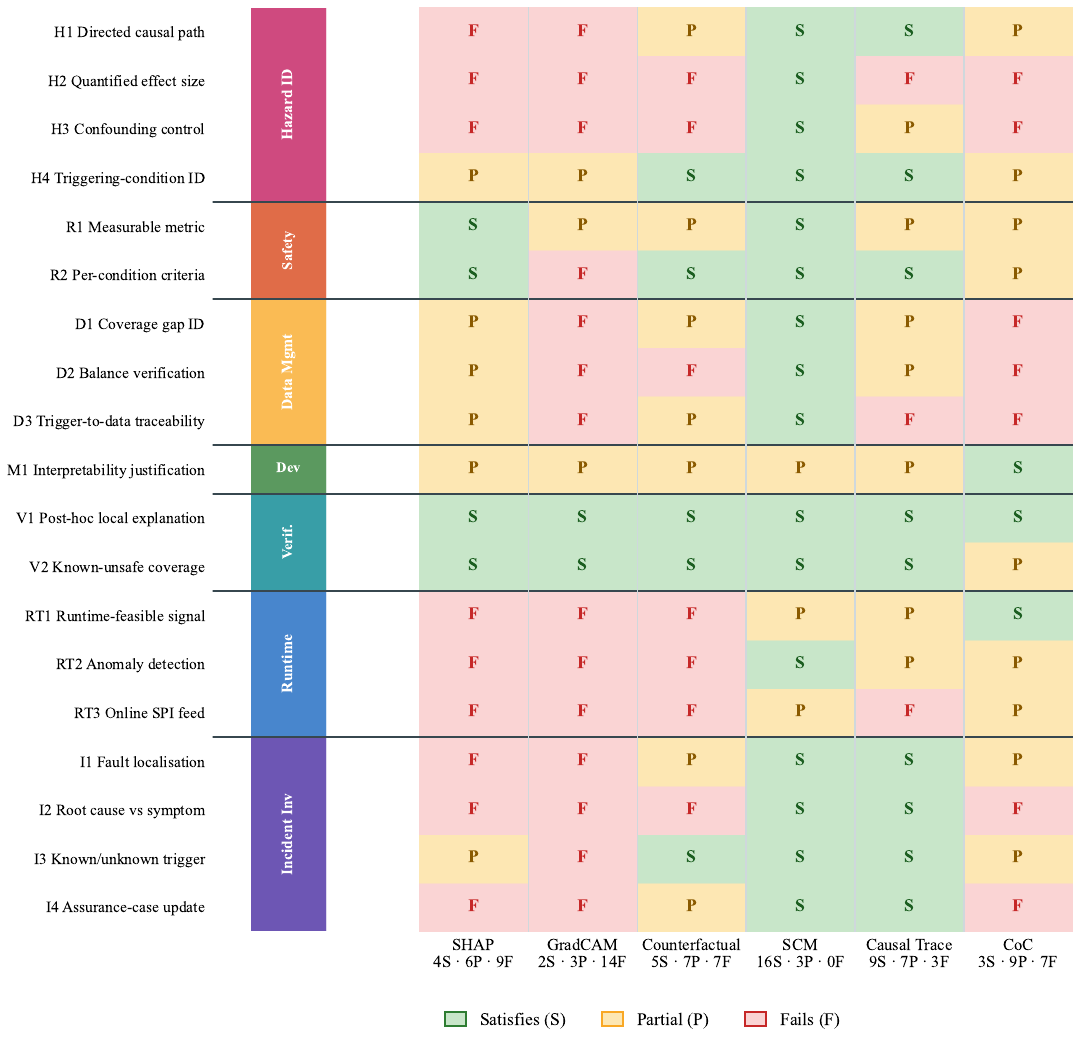}
\caption{\textbf{SCM is the only column with zero structural failures across the $6 \times 19$ rubric.} Rows: criteria grouped by lifecycle stage; columns: methods with column totals $S\!\cdot\!P\!\cdot\!F$. Green cells satisfy the criterion structurally, yellow partial, red fail.}
\label{fig:rubric_heatmap}
\end{figure}

Column totals in Fig.~\ref{fig:rubric_heatmap} give the global coverage ranking: SCM~$16\mathrm{S}\cdot3\mathrm{P}\cdot0\mathrm{F}$ (the only zero-failure method); Causal Trace $9\mathrm{S}\cdot7\mathrm{P}\cdot3\mathrm{F}$; Counterfactual $5\mathrm{S}\cdot7\mathrm{P}\cdot7\mathrm{F}$; SHAP $4\mathrm{S}\cdot6\mathrm{P}\cdot9\mathrm{F}$; CoC Trace $3\mathrm{S}\cdot9\mathrm{P}\cdot7\mathrm{F}$; GradCAM $2\mathrm{S}\cdot3\mathrm{P}\cdot14\mathrm{F}$.

\subsection{Stage-Level Verdicts}

\begin{figure}[t]
\centering
\includegraphics[width=0.85\columnwidth]{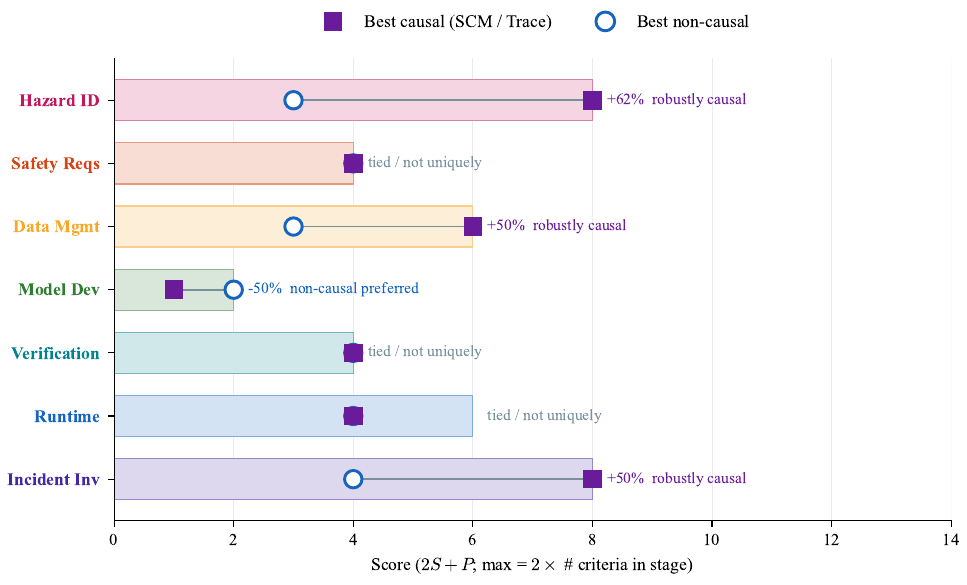}
\caption{\textbf{Three stages clear the $30\%$ causal-necessity threshold; all survive a one-step cell flip.} Best causal-method score (filled square) vs best non-causal (open circle) per stage, with the verdict annotation. Hazard ID survives any one-step flip, incident investigation survives any one-step flip, and data management has a 3-criterion margin (D1, D2, D3 each favour SCM).}
\label{fig:rq3_verdict}
\end{figure}

The $\Delta \geq 30\%$ rule is conservative: at 30\%, a verdict requires that causal methods outperform all non-causal alternatives by at least two full S-vs-F flips on a 4-criterion stage. Three verdicts qualify (Fig.~\ref{fig:rq3_verdict}): hazard identification ($+62\%$), incident investigation ($+50\%$), and data management ($+50\%$ across D1/D2/D3). The verdict set is stable across threshold choice: the same three stages qualify for any threshold $T \in (0\%, 50\%]$, and no additional stage qualifies at any positive threshold (the four non-qualifying stages tie at Safety Requirements, Verification, and Runtime, or favour CoC at Model Development on raw scores).

All three verdicts survive any single one-step cell flip (one notch, $\mathrm{S}\!\to\!\mathrm{P}$ or $\mathrm{P}\!\to\!\mathrm{F}$), with worst-case post-flip gaps of $+50\%$ (hazard ID), $+37.5\%$ (incident investigation), and $+33\%$ (data management, where SCM is \textbf{S} on all three D-criteria and no non-causal method is \textbf{S} on any), each above the $30\%$ threshold. Evidential strength ranks hazard ID $\succ$ incident investigation $\succ$ data management by direct-clause coverage. The S on H2, I2 and similar cells is a claim about output \emph{type}; fidelity of the specific quantities produced by our fit is addressed empirically in Section~\ref{sec:validation}.

\subsection{Limitations of the Structural Analysis}

The rubric evaluates output types, not output quality. An SCM with incorrect edges still scores S on H1. Section~\ref{sec:validation} addresses this empirically for one VLA. Three of six methods are scored structurally only; GradCAM is architecturally inapplicable to the VLA under test, so its column rests on method-class reasoning rather than a fit. Among the three empirically touched methods, D1 is illustrated by the learned graph's sparsity pattern; D2 and D3 rest on structural scoring alone in this paper.

%% ─────────────────────────────────────────────────────────────────
\section{Proof-of-Concept Validation}
\label{sec:validation}
\subsection{Dataset and Setup}

\begin{figure}[tb]
\centering
\includegraphics[width=0.85\columnwidth]{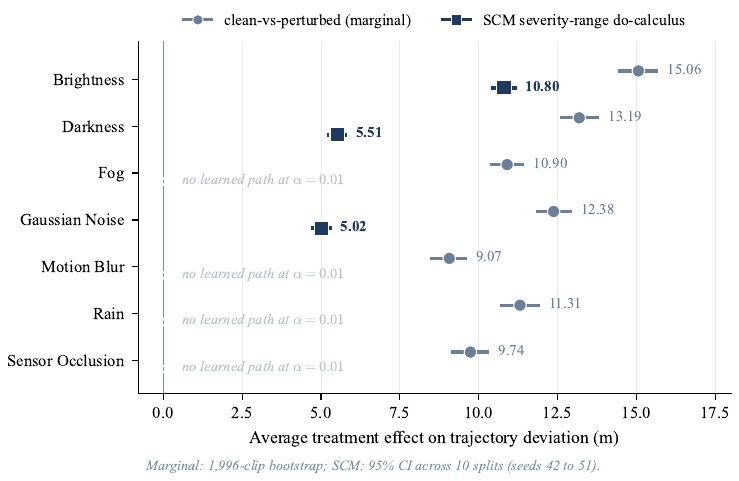}
\caption{\textbf{The largest empirical gap in the paper: marginal effect estimates recover all seven perturbation types; SCM do-calculus recovers only three.} \emph{Clean-vs-perturbed} (marginal, clip-level bootstrap over all $1{,}996$ clips) recovers all seven types. \emph{SCM severity-range do-calculus} ($\operatorname{do}(\mathrm{sev}{=}0)$ vs $\operatorname{do}(\mathrm{sev}{=}1)$ on the normalised per-type severity axis, 10 splits) recovers only three; four never enter the graph at $\alpha=0.01$. The structural rubric predicts all seven \emph{can} be recovered by an SCM; our fitted instance does not.}
\label{fig:ate_forest}
\end{figure}

The structural rubric (Section~\ref{sec:gap}) establishes which output \emph{types} are admissible at each lifecycle stage; it does not establish that any specific instance of an admissible method will produce content correct enough for compliance. Structural admissibility is necessary but not sufficient. This section tests the necessary half: do the three methods we run on a production VLA produce the output types the rubric predicted? Empirical sufficiency (whether a fitted SCM's edges, a learned anomaly's localisation, or a trace's named cause are themselves correct) is the open question we return to in Section~\ref{sec:discussion}. We evaluate three representative methods (SHAP at rung~1, a structural causal model at rung~2/3, and a Chain-of-Causation language trace) on $1{,}996$ real-world driving clips from the NVIDIA PhysicalAI-AV dataset~\cite{PhysicalAI}, under seven perturbation conditions at multiple severities ($79{,}840$ rows). A shared clip-level $80/20$ split is repeated across ten seeds (42 to 51); $95\%$ CIs are reported for all three aggregates. Scope caveats: SHAP is fit on perturbed-only rows; the SCM learns on the full split (including $1{,}996$ clean rows); CoC rates aggregate pre-computed \texttt{*\_changed} flags, not a fresh inference run. The SCM ATE in Fig.~\ref{fig:ate_forest} is a severity-range effect on the normalised severity axis; a complementary clean-vs-perturbed ATE is reported in the caption. GradCAM is architecturally inapplicable to this VLA (Section~\ref{sec:gap}). Effect estimates use clip-level bootstrap ($n = 1{,}996$ clips) for marginal aggregates; method-level scoring metrics ($R^2$, edge counts, change rates) aggregate over all $79{,}840$ rows ($1{,}996$ clean plus $77{,}844$ perturbed).

\subsection{Method-Level Scoring Verification}

For each method, we check whether the output observed on this VLA matches the output type the rubric predicts. \textbf{SHAP} (rubric: F/F/F on H1/D1/I1) produces a ranked feature list: mean test $R^{2}=0.119$ $[0.100, 0.139]$ on trajectory deviation across ten splits (target clipped at train-only $q_{99}$); the top-ranked feature is a generic severity score; local SHAP recovers the active perturbation in its top-5 for $52.9\%$ of matched silent-failure cases. The output contains no directed edges (H1: F), no distribution-level signal (D1: F), and no named root-cause variable (I1: F), as predicted. \textbf{SCM} (rubric: S/S/S) produces a DAG with mean $6.7$ edges $[6.1, 7.3]$ across ten splits (six core edges stable in 10/10). Three perturbation types (brightness, darkness, Gaussian noise) reach trajectory deviation in 10/10 splits; four (fog, motion blur, rain, sensor occlusion) never enter the graph at $\alpha=0.01$. The output is directed (H1: S), sparse-for-4/7-types (D1: S, with the honest caveat that absence may reflect real weak effects rather than a data gap), and node-attributable (I1: S by output type; empirical fidelity is the open challenge below). \textbf{CoC trace} (rubric: F/F/P) shows per-perturbation $\mathrm{action\_changed}$ rates of $10.1\%$ $[9.5, 10.6]$ (motion blur) to $21.0\%$ $[20.4, 21.7]$ (brightness); CoC consistency correlates with trajectory safety at $r=0.99$~\cite{priyadershi2026}. No directed graph (H1: F), no distribution signal (D1: F), natural-language cause candidates rather than formal attribution (I1: P).

On this single VLA, each method's observed output type matched its rubric prediction (a sanity check, not a validation): output types were predicted a priori from method descriptions, so observed consistency is descriptive of the rubric, not corroborative. The empirical findings worth carrying forward are the silent-failure example below (active perturbation ranked 8th by local SHAP) and the 4-of-7 missed SCM edges; together they expose the next assurance challenge: validating causal-explanation \emph{fidelity}, not just output type. Three of six method classes (GradCAM, counterfactual, causal tracing) are scored structurally only in this paper.

\paragraph{One representative silent failure.}
Consider a low-speed stop-signal approach under a brightness perturbation, with trajectory deviation $32.37$~m against $q_{75}{=}14.76$~m. SHAP's top-5 contains four trajectory-shape features and ranks the active perturbation eighth; the SCM reads the trigger via metadata but its learned anomaly points to a downstream node (a symptom); CoC fields are unchanged. Since \emph{silent failure} is defined here as ``CoC unchanged \textsc{and} endpoint ${>}1$~m,'' the CoC observation on this row is part of the subset definition, not an independent evaluation. The non-tautological check: on $19{,}461$ perturbed rows with trajectory deviation above $q_{75}$ (no CoC conditioning), a CoC field changes in $48.4\%$, consistent with CoC as a partial safety signal.

\paragraph{Structural-discovery sensitivity.}
Re-running PC+GES at $\alpha \in \{0.01, 0.05, 0.10\}$: brightness, darkness, Gaussian noise 10/10 at every $\alpha$; fog 10/10 at $\alpha{\ge}0.05$; motion blur, rain, sensor occlusion 0/10 at every $\alpha$. Four of seven recover at the standard $\alpha{=}0.05$. For the remaining three, within-type dependence measures confirm the effect exists (comparable in magnitude to recovered types); PC+Fisher-Z fails on the full 79{,}840-row matrix because these three combine smaller effect magnitudes with sparser column representation ($\ge$92\% zeros). This is a power/encoding limit of one CI test, not effect absence; a per-type stratified fit or KCI is a natural next step. The remaining open questions are consolidated in Section~\ref{sec:discussion}.

%% ─────────────────────────────────────────────────────────────────
\section{Discussion}
\label{sec:discussion}
\subsection{Implications for XAI Method Selection}

The \emph{evidence-type gap} has a direct consequence for ISO/PAS~8800 and ISO~21448 compliance: selecting an XAI method by popularity risks leaving stage-specific evidentiary gaps unclosed. Fig.~\ref{fig:rubric_heatmap} gives the per-stage picture. SHAP is admissible at safety requirements and verification; GradCAM only at verification. CoC and SCM tie at 4/6 on runtime. This admissibility profile aligns with current practice, where SHAP, GradCAM, and attention-based methods already dominate Model Development and Verification (Section~\ref{sec:usage}); the paper's contribution lies at the three lifecycle stages where current practice does not yet meet the evidence type the standards require. Among causal methods, only SCM is all-\textbf{S} on the three causal-necessary stages; causal tracing scores \textbf{F} on H2 and D3, so is not all-\textbf{S}. Prior XAI-to-safety mappings organised by ADS task~\cite{kuznietsov2024}, technical approach~\cite{atakishiyev2024}, or practitioner need~\cite{fresz2024} are complementary; none derives testable criteria from clause text and scores methods structurally.

\paragraph{Where the pilot lands.}
Section~\ref{sec:validation} reports single-VLA findings: directed paths + stable ATEs (10/10 splits) for three perturbation types at $\alpha{=}0.01$; an $\alpha$-sensitivity sweep recovers a fourth (fog) at $\alpha{=}0.05$, with three (motion blur, rain, sensor occlusion) remaining undetected at $\alpha{=}0.10$. Those three do show non-zero clean-vs-perturbed marginal effects, so the missed edges indicate a limit of the Fisher-Z CI test on this dataset rather than absence of effect. Learned root-cause diagnosis from downstream signals is at chance ($30.4\%$ vs.\ $30.8\%$ majority) in our benchmark, which is the central open challenge.

\paragraph{Correlational evidence as a proxy.}
A method that is inadmissible for a stage's evidence \emph{type} is not thereby without practical use at that stage. Because admissibility is necessary but not sufficient, a structurally admissible causal model with weak fitted content (here, the SCM recovering only three of seven perturbation types) can be outperformed in practice by a correlational screen: clean-vs-perturbed marginal effects recover all seven types, and local SHAP surfaces the active perturbation in its top-5 for $52.9\%$ of matched silent-failure cases (Section~\ref{sec:validation}). A correlational method can therefore act as a proxy or screening signal (surfacing candidate triggers, prioritising investigation, flagging anomalies), particularly where the available causal alternative is empirically weak, even though it provides association and not the cause-and-effect chain the standards require. This is the type-versus-quality distinction applied to an inadmissible type: such a method may still supply useful supporting evidence, just as an admissible type does not guarantee sufficient content, provided it is not treated as the primary causal evidence the assurance case must contain.

\paragraph{Limitations.}
\textbf{Rubric validity:} self-scored; mitigated by structural (output-type) scoring, single-flip verdict robustness, and the validation protocol below. \textbf{Scope:} three of six methods scored structurally only; empirical pilot is single-dataset, single-VLA. \textbf{Interpretive criteria:} H2, H3, I2, D2, D3 are clause-formalisations; data-management verdict direction is robust to any single-cell flip, but magnitude depends on our reading of Cl.~11.3.6/.7. \textbf{Literature map:} Section~\ref{sec:usage}'s distribution is inferred from two surveys' categorisations, not a re-coding of primary papers.

\subsection{Future Work}

(i)~\textbf{Learned diagnosis from downstream signals} via abductive causal inference (our baseline is at chance). (ii)~\textbf{Cross-dataset/model replication} on STRIDE, nuScenes, other VLAs. (iii)~\textbf{Two-panel rubric validation}: construct-validity on the interpretive criteria (H2, H3, I2, D2, D3) and method-scoring on the full $19 \times 6$ matrix, with 2--3 external raters, blind scoring then adjudication, weighted $\kappa$/Krippendorff's $\alpha$ pre- and post-adjudication, and verdicts recomputed per rater. We pre-commit to downgrading any contested verdict.

\section{Conclusion}

% SASSUR framing: emphasise assurance-case / certification deliverable.
XAI papers in ADS are organised by output type and technique family; safety standards are organised by evidentiary demand. Bridging the two requires reading the clauses literally and asking, of each method, whether its output \emph{type} contains what a clause requires. Our 19-criterion rubric, derived from four publications with representative clause-cited derivations and a reproducible scoring rule, makes that question answerable, and produces three robust verdicts: causal XAI is the only class that clears our $\geq 30\%$ causal-necessity threshold at hazard identification, incident investigation, and data management, under our standards-derived interpretation; at the remaining four stages, correlational or language-based methods are also admissible. The rubric and its scoring rule are domain-general; we offer them as shared evaluation infrastructure for XAI method selection in ADS safety assurance, where selection by method popularity should be replaced by selection by lifecycle-stage demand.

%% ─────────────────────────────────────────────────────────────────
\begin{credits}
\subsubsection{\discintname} All authors are NVIDIA employees. The empirical pilot (Section~\ref{sec:validation}) uses NVIDIA's PhysicalAI-AV dataset and one NVIDIA vision-language-action system; the rubric and verdicts (Sections~\ref{sec:standards}--\ref{sec:gap}) are independent of any NVIDIA product. No external funding was received.
\end{credits}

%% ─────────────────────────────────────────────────────────────────
\bibliographystyle{splncs04}
\bibliography{references}

\end{document}